# A Long-term Dependent and Trustworthy Approach to Reactor Accident Prognosis based on Temporal Fusion Transformer

Chengyuan Li[1], Zhifang Qiu[1], Yugao Ma[1], Meifu Li[1, *]

*Corresponding author. Email: meifu_lee@163.com. Tel: (+86)13880702386

[1]Science and Technology on Reactor System Design Technology Laboratory, Nuclear Power Institute of China, Chengdu, 610213, China

## Abstract

Prognosis of the reactor accident is a crucial way to ensure appropriate strategies are adopted to avoid radioactive releases. However, there is very limited research in the field of nuclear industry. In this paper, we propose a method for accident prognosis based on the Temporal Fusion Transformer (TFT) model with multi-headed self-attention and gating mechanisms. The method utilizes multiple covariates to improve prediction accuracy on the one hand, and quantile regression methods for uncertainty assessment on the other. The method proposed in this paper is applied to the prognosis after loss of coolant accidents (LOCAs) in HPR1000 reactor. Extensive experimental results show that the method surpasses novel deep learning-based prediction methods in terms of prediction accuracy and confidence. Furthermore, the interference experiments with different signal-to-noise ratios and the ablation experiments for static covariates further illustrate that the robustness comes from the ability to extract the features of static and historical covariates. In summary, this work for the first time applies the novel composite deep learning model TFT to the prognosis of key parameters after a reactor accident, and makes a positive contribution to the establishment of a more intelligent and staff-light maintenance method for reactor systems.

**Keywords:** Temporal Fusion Transformer; Prognosis; Loss of Coolant Accident; Multi-horizon Forecasting

## 1. Introduction

Mitigating climate change by reducing greenhouse gas emissions has been a central challenge worldwide since the 21st century. Nuclear energy, a low-carbon energy source just below hydropower in terms of global installed capacity, accounts for more than a quarter of global low-carbon electricity supply each year and 10% of global electricity generation. Therefore, in order to achieve the goal of a global average temperature rise within 2°C by 2050, it is necessary to continue to promote the construction of Gen III nuclear power plants[1].

At present, Gen III nuclear power technology is dominated by advanced pressurized water reactor technology, which is the preferred technology route for new nuclear power units. Although Gen III nuclear technology is capable of controlling the frequency of core damage to less than $1.0\times10^{-5}$ per reactor year, there is still the potential for a large release of radioactive material, i.e., a working condition known as an "Extreme Accident (EA)" [2].

The EA condition consists of a variety of typical initiating events, as shown in Figure 1. Among the possible initiating events of an EA, the probability of a loss of water accident (LOCA) is rare, but since the consequences are quite serious, the acceptance guidelines for EA operating conditions in the Safety Analysis Report (SAR) are based on the LOCA analysis[3].

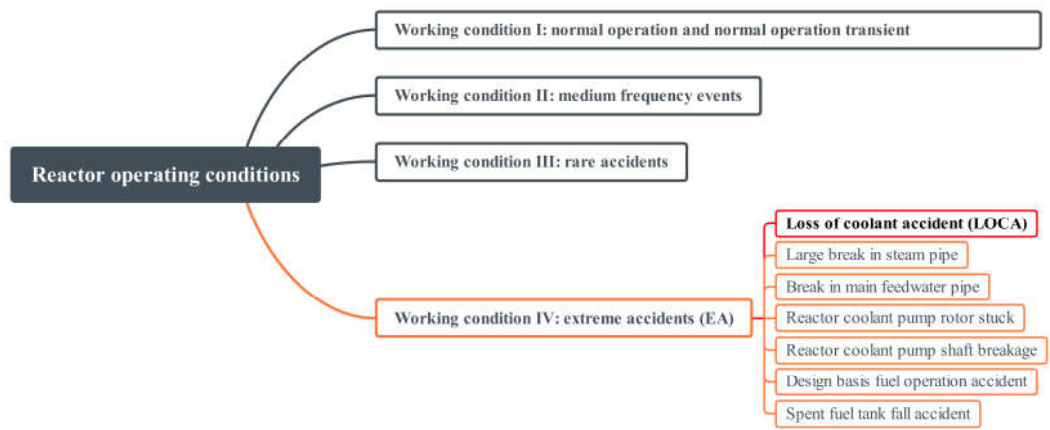

Figure 1 Composition of reactor operating states and extreme accident conditions

In order to ensure core safety, i.e., no core meltdown, even after a LOCA, a series of proactive mitigation measures are required after the occurrence of an incident. For the Gen III nuclear power technology HPR1000, accident management relies on sign-oriented protocols, which can be further divided into Optimal Recovery Protocols (ORPs) and Functional Recovery Protocols (FRPs) depending on the design approach. If the cause of the accident can be diagnosed and is part of certain accidents, ORP is used, but this is only for a few conditions, such as large breaks with an equivalent diameter greater than 34.5 cm; if the accident cannot be diagnosed or the accident does not correspond to the specific accidents for which ORP is used, which corresponds to most of the accident conditions, the reactor needs to be disposed of using FRP, such as most middle breaks with an equivalent diameter in the range of 2.5~34.5cm. Therefore, diagnosing and prognosing accidents after FRP is activated is extremely crucial for operators to take appropriate disposal decisions and prevent further deterioration of the accident status. There are two key steps in the disposal process: first, the correct diagnosis of the accident can provide interface data for the prognosis of the reactor status; then, the correct prognosis of the reactor status change can provide effective reference for the operator's decision. The work in this paper focuses on post-LOCA prognosis of the reactor, while using the results obtained in the previous phase, i.e., accident diagnosis.

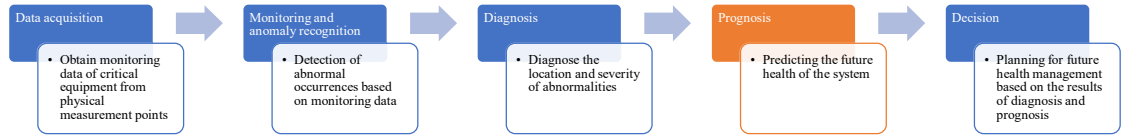

Figure 2 Prognosis in the sequence of reactor accident health maintenance

## 1.1 Literature review

A considerable number of previous explorations have been carried out in order to rapidly diagnose reactor anomalies and to foresee the process of anomalies. Most of these methods are model-based or data-based, or rule-based methods built on models and data combined with expert knowledge [4–6].Currently, data-based approaches based on statistical learning and deep learning are the focus of research, due to their greater generalization ability and inference speed compared to model-based approaches, and their easier maintenance than rule-based approaches that utilize large knowledge bases.

The diagnosis of abnormal reactor operation has been explored extensively by previous works. Lin et al [7] developed a Nearly Autonomous Management and Control (NAMAC) framework for advanced reactors and used feedforward neural networks for the digital twin (DT) layer in the framework to quickly identify information about anomalous transients. Ayodeji et al [8] constructed a nuclear power plant operator operation support system based on the use of principal component analysis (PCA) and two different neural networks: an Elman-type recurrent neural network (Elman-RNN) and a radial basis neural network (RBN) for fault diagnosis. Lee et al [9] organized the large amount of real-time data generated by a single system and the dynamics of the individual system monitoring data by constructing two-channel 2D images and used convolutional neural networks (CNN) for feature extraction and diagnostic tasks of system anomalous transients. Wang et al [10] proposed a support vector machine (SVM)-based diagnosis method in order to improve the diagnostic capability of the model on a smaller number of accident instances and used an improved particle swarm optimization (PSO) method for the selection of hyperparameters of the model to achieve an improved accident classification capability in the case of small samples. Li et al [11] constructed an integrated learning model using various statistical learning models and neural network models, such as SVM, random forest model (RF), k-nearest neighbor model (KNN), and fully connected neural network (FCNN), and based on multivariate voting method and weighted voting method, the model was able to achieve a rapid response and robust to noise for accident diagnosis. For more detailed information on diagnostic methods for abnormal reactor operation, please refer to the review articles [12–15].

Compared to the task of accident diagnosis, previous research done in post-accident prognosis is relatively insufficient, and this is especially true for the work on prognosis of system-level parameters. Although the use of best estimation (BE) based system analysis programs, such as RELAP[16] or ARSAC [17], allows for a more accurate calculation of accidents with known parameters, their use in real-time operator manipulation is still impractical due to their slow computational speed. Therefore, the use of data-based pre-trained models for fast inference calculation is an important alternative method to have the ability to perform ultra-real-time inference of the reactor state after an accident. In for the task of system state prediction after the occurrence of anomalous reactor transients, Zeng et al [18] used SVM to construct agent models for the thermal and physical steps of nuclear thermal coupling calculations, respectively, along with a particle filtering framework for noise filtering and prediction of system parameter measurements to achieve the system state prediction task for the Transportable Fluoride-salt-cooled High-temperature Reactor (TFHR) under reactive introduction accidents. Koo et al [19] used FCNN to construct a model for predicting the trend of pressure vessel (PV) water level for steam generator pipe rupture and cold/hot leg LOCA and demonstrated the superiority of this prediction method by comparing the performance of this model with a cascaded fuzzy logic neural network model (CFNN) for the same task. Zhang et al [20] adopted a long short-term memory network (LSTM) that is more sensitive to less data in order to improve the quantitative imbalance between the training data on the fluctuating operation category and the stable operation category, and used the model trained using this strategy for the task of predicting the reactor's pressurizer (PRZ) water level under abnormal operation. Gurgen et al [21] developed a physically constrained LSTM reactor parameter prediction method based on physical constraints in order to serve the decision layer in the NAMAC framework, and applied the method to the prediction of fuel centerline temperature in the loss-of-flow accident (LOFA) of Experimental Breeder Reactor II. However, with the exception of the few studies on reactor state prediction at the system level mentioned above, most prediction efforts have focused more on the state and remaining usable time (RUL) of subsystems or components. For example, Ramuhalli et al [22] used LSTM model, SVM model and nonlinear autoregressive model (NAR) to predict the operating status of feedwater and condensate system (FWCS) of boiling water reactor (BWR) for the next day and the next week, respectively. Liu et al [23] proposed a dynamic weight integration learning prediction method based on multiple SVM regression models, and applied the method to the leakage prediction task of the reactor first-loop coolant main pump, as well as estimated for the uncertainty of the prediction results. More RUL prognostic tasks on reactor subsystems and components are available in the review articles by Ayo-Imoru et al [24] and Si et al [25].

Unlike the nuclear field, prognostic tasks are gaining importance in many other industries. Most of these methods are data-based. In the new energy industry, state prediction for battery packs after anomalies is of great importance for the safety of electric vehicles. Hong et al [26] established an accurate multi-forward-step voltage prediction method for battery systems using LSTM and validated the superiority, stability and robustness of the method using real-world data. Liu et al [27] proposed a joint prognostic method of AutoRegressive Integrated Moving Average model (ARIMA) and LSTM using approximate optimization method in order to improve the prediction accuracy of electric vehicle battery pack voltage. In the field of wind power generation, prediction of the degradation process of turbines and estimation of RUL is an essential means to improve the operating economics [28]. Saidi et al [29] constructed a prediction method based on spectral kurtosis and SVM regression model in order to predict the operating condition of the high-speed shaft bearing in wind turbines. Encalada-Dávila et al [30] constructed a method for predicting Low-speed shaft temperature under normal operation and abnormal transients using fully connected neural networks and validated the method on real-world operational data from several wind turbines. Therefore, the prognostic approach in the non-nuclear field can provide insights and inspiration for the task of prognosis of system level parameters after a reactor accident.

Deep learning theory is providing novel solutions to traditional problems in various industries thanks to the ability of neural networks using gradient descent as an optimization method to act as a function fitter for any data type, and the ability of this fitter to automatically capture the effective features of the original data. For the task of predicting the accident process, modeling of temporal data is required, so most of the previous work has used recurrent neural networks (RNNs), which incorporate Elman-type RNNs, long short-term memory (LSTM), and gated neural units (GRU) to increase the accuracy of predicting the relevant parameters. However, RNNs forget early inputs in long-sequence scenarios, i.e., the long-range dependence problem; and the model itself has difficulty in estimating the prediction uncertainty. Meanwhile, the existing accident prediction methods do not fully use the results of accident diagnosis and the variation of other parameters outside the target parameters, which reduces the efficiency of data utilization.

## 1.2   Contributions of the work

In this study, a novel method for predicting important parameters after a reactor accident is proposed in order to remedy the long-term dependence problem, lack of uncertainty estimation, and inefficient data utilization in previous work on reactor accident parameter prediction. Thus, this paper attempts to make three significant contributions and improvements to the current technique as follows:

1）    A Temporal Fusion Transformer (TFT) model that improves the RNN long-range dependency problem is developed. This model not only models the temporal data using classical RNN, but also utilizes the state-of-the-art Transformer architecture in computer natural language processing (NLP) in order to automatically capture the remote associations of elements in long-range sequences.

2）    A prediction uncertainty estimation method based on stochastic processes is developed. Reactor accident management belongs to a scenario with high safety requirements, and the estimation of prediction intervals can produce best and worst-case indications of target parameters, which can help optimize subsequent accident management decisions. The parameters fitted by the TFT model used in this paper during the learning process are the distribution information of the parameters at each time stamp, so the sampling information of the joint distribution of all prediction steps is obtained by Monte Carlo (MC) sampling during the prediction process.

3）    A prediction method with perception of accident diagnostic labels and multiple monitored parameters is developed. Multiple other monitorable thermal parameters can be used as historical covariates in the prediction of the target parameters, and diagnostic labels for the type and severity of the upstream accident are supported as static covariates. This prediction method using multiple covariates has been validated to improve prediction accuracy and increase the efficiency of data usage.

## 1.3   Organization of the paper

The structure of the remainder of this paper is organized as follows: the second part describes the prediction methods and related technical details for key parameters after a reactor accident; the third part presents information about the data set and the model parameters selected for the experiments; the fourth part details the experimental procedure and its results, and analyzes the results; the fifth part concludes the work of this paper.

## 2. Methodology

### 2.1 Task description

The task of this study is the prediction of key parameters of the reactor after LOCA, which is a multi-horizon prediction problem in which the variables of interest are predicted over multiple future time steps. For any parameter $y$ to be predicted, the variables that can assist in the prediction are the static covariates $s \in \mathbb{R}^{m_s}$ of the upstream incident diagnosis, the known scalars $y_t \in \mathbb{R}$ of the parameter $y$ at each time step on the historical timeline $t \in [0,T]$, and the sampled values $\chi_t \in \mathbb{R}^{m_\chi}$ of the other covariates at the time step that assist in the prediction. The other covariates contain two components, the historical covariates and the future covariates $\chi_t = \left[z_t^\mathrm{T}, x_t^\mathrm{T}\right]^\mathrm{T}$, where $z_t \in \mathbb{R}^{m_z}$ are time-series parameters that are not known after the forecast moment, and $x_t \in \mathbb{R}^{m_x}$ are predictable after the start of the forecast, such as information on the duration of the accident.

In order to know the confidence interval for each prediction step in the prediction, TFT uses quantile regression methods to make inference for each of the 10%, 50% and 90% quantile points at any moment. Thus, for any parameter to be predicted the prediction problem can be expressed as

$$\hat{y}(q,t,\tau) = f_q\left(\tau, y_{t-k:t}, z_{t-k:t}, x_{t-k:t+\tau}, s\right) \tag{1}$$

where $q$ is the percentile of the point to be predicted; $t$ is the starting moment of the prediction task; $\tau \in \mathbb{Z}^+$ is the distance between the point to be predicted and the starting point of the prediction; $k$ is the size of the time window utilized for the forecasts; the corner label $t-k:t$ contains $k+1$ elements. The description of the multi-horizon prediction problem after LOCA is shown in Figure 3.

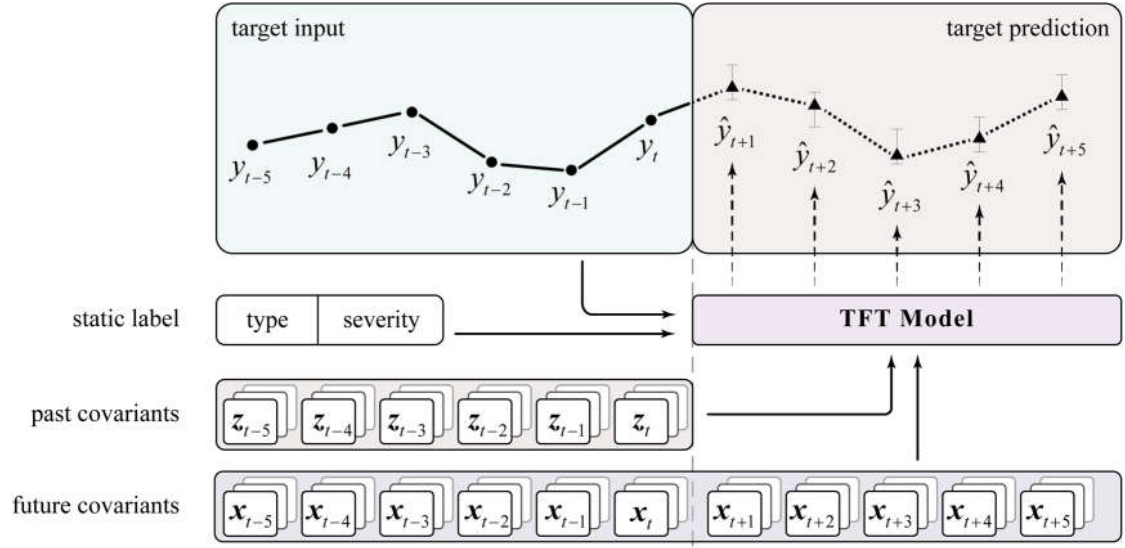

Figure 3 The visual description of the multi-horizon prediction problem for post-LOCA

## 2.2 Temporal Fusion Transformer for LOCA prognosis

### 2.2.1 Overview

Temporal Fusion Transformers (TFT) is a novel deep neural network-based time series forecasting framework, most notably characterized by the introduction of a self-attention mechanism and the ability to provide interpretability of model prediction results in specific cases. This prediction framework achieves superior performance over commonly used model benchmarks on open source datasets, which include the classical ARIMA, LSTM, GRU, but also includes DeepAR, MQRNN, ConvTrans, etc., which have complex network structures[31]. Overall, in the task of parameter prediction after LOCA of the reactor, TFT possesses five main technical features that ensure its excellent performance on the multi-horizon prediction problem, namely: 1) Use of gating mechanism: In the process of traversing the visible temporal dynamics, the gating mechanism can filter out irrelevant data points to reduce their interference with the prediction. 2) Input variable embedding module: In each time step, this module is able to embed a tiled vector of multiple covariates and known target variables into a fixed dimensional vector to facilitate the transfer of data in the model. 3) Static covariate encoder: this session will make full use of the accident diagnosis information, converting accident category labels and scalar labels into conditional information that constrains the calculation of each prediction step. 4) Dual temporal dynamic processing: On the one hand, LSTM is used to process the temporal data using sequence-to-sequence (Seq2Seq) approach to capture the data features of the data under short period; on the other hand, the distant relationships and features of the temporal data are captured using the Transformer layer based on the multi-headed self-attention mechanism, which thoroughly improves the long-term dependency problem in the traditional Seq2Seq. 5) Forecast range estimation: The quantile regression forecasting method is used to determine the possible range of target parameter values for each forecast time step. The flow of the prediction using the TFT model is shown in Figure 4.

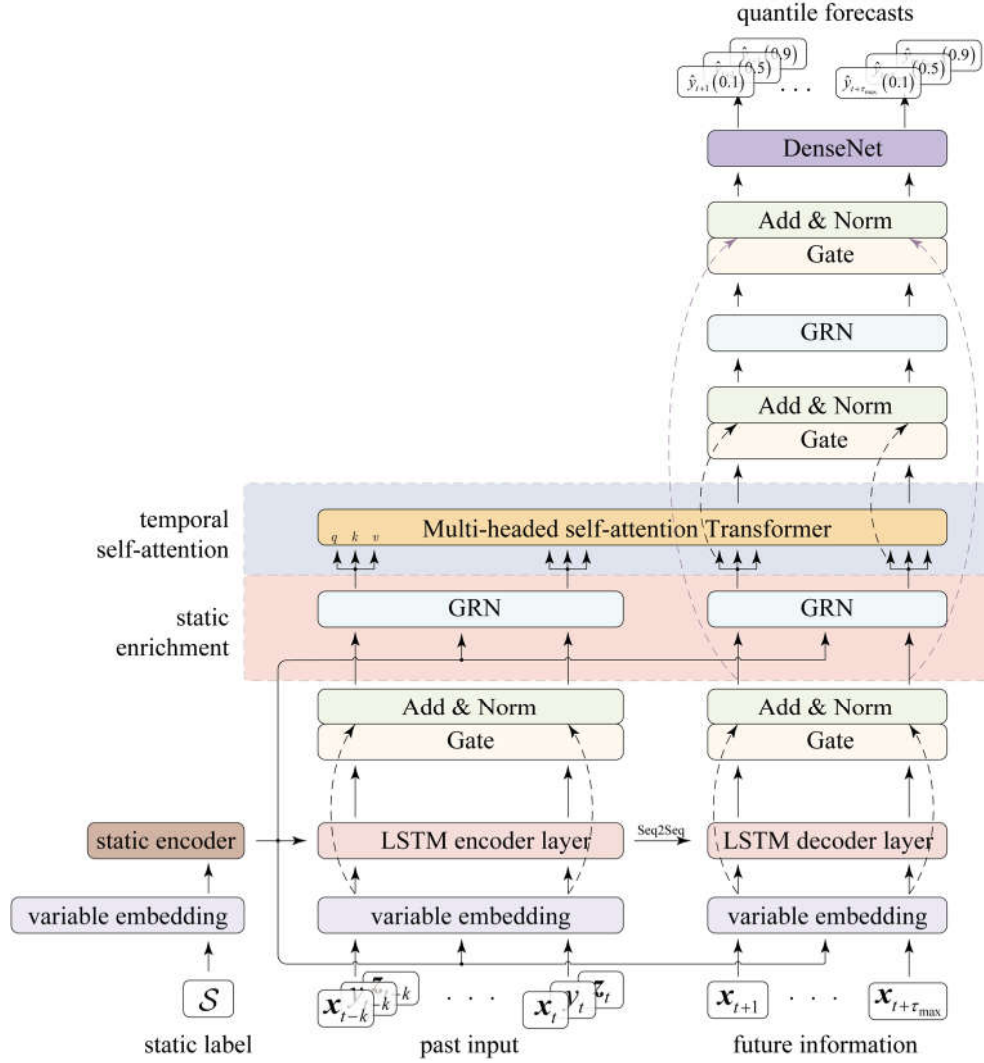

Figure 4 A framework for multi-covariate prediction using TFT models

### 2.2.2 Gating mechanisms

Since the relationship between the input time series data for the output predicted values is not known at the beginning, a gated residual network (GRN) can be used to give more flexibility to the prediction model. This can filter out the data with significant relationships and feed them into the network. GRN reads an embedding vector $a \in \mathbb{R}^{d_{model}}$ of several reactor monitoring data in one timestamp at a time, and optionally a vector $c \in \mathbb{R}^{d_{model}}$ associated with the input conditions, and the output is a vector that passes through the same dimensions as the embedded vector. The GRN is calculated as follows

$$\begin{aligned}
\text{GRN}(a,c) &= \text{LayerNorm}(a + \text{GLU}(\eta_1)), \\
\text{GLU}(\eta_1) &= \sigma(W_4\eta_1 + b_4) \odot (W_5\eta_1 + b_5) \\
\eta_1 &= W_1\eta_2 + b_1, \\
\eta_2 &= \text{ELU}(W_2 a + W_3 c + b_2)
\end{aligned} \quad (2)$$

where $W_{(\cdot)} \in \mathbb{R}^{d_{model} \times d_{model}}$ is the weight of the connection to be learned; $b_{(\cdot)} \in \mathbb{R}^{d_{model}}$ is the bias to be learned; $\odot$ is the element-wise Hadamard product; $\sigma(\cdot)$ is the sigmoid activation function; $\mathrm{ELU}(\cdot)$ is the activation function with soft saturation on the left side and no saturation on the right side; $\mathrm{LayerNorm}(\cdot)$ is the standard layer normalization method [32]; $\mathrm{GLU}(\cdot)$ is the gate component of the Figure 5. The computational flow of GRN is shown in Figure 5.

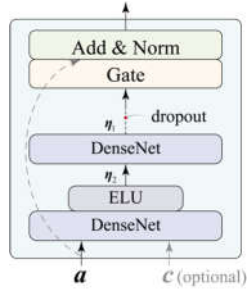

Figure 5 Calculation flow of GRN module

### 2.2.3 Variable embedding module

Because of the difference in the number of historical covariates as well as future covariates, the vector dimensions of the input model are different at each time step in history and in the future. Also, static covariates usually have different dimensions from the previous two covariates. In order to ensure dimensional consistency of the data inputs for subsequent models, it is necessary to embed the data dimensions of the three covariates at each time step. For any moment on the time axis, the covariates can be expressed as $\theta_t \in \mathbb{R}^{d_{in}}$, and satisfy $d_{in} \in \{m_s, m_x + m_z, m_x\}$. The individual elements are first expanded into a vector $\Theta_t = \theta_t W_{mlp} \in \mathbb{R}^{d_{in} \times d_{model}}$ of dimension $d_{model}$, where $W_{mlp} \in \mathbb{R}^{d_{model}}$ is a learnable variable, and the weights are used in all time steps. Subsequently, each column vector of $\Theta_t$ is passed through a GRN with network weight $\omega_1$, respectively, and the individual column vectors of the output are combined according to the original arrangement to obtain $\Theta_{t,\text{gated}} \in \mathbb{R}^{d_{in} \times d_{model}}$. Meanwhile, the individual column vectors of $\Theta_t$, together with the optional conditional vector $c$, are simultaneously passed through a GRN with network weight $\omega_2$ and then passed through the Softmax layer to obtain the parameter weight vector $w_t = \mathrm{Softmax}(\mathrm{GRN}_{\omega_2}(\Theta_t, c)) \in \mathbb{R}^{d_{in}}$. Finally, the individual column vectors of $\Theta_{t,\text{gated}}$ are weighted and summed with the elements corresponding to the parameter weight vectors to obtain the final covariate information of the input TFT model at that time step, i.e., $\tilde{\theta}_t = \Theta_{t,\text{gated}}^T v_t \in \mathbb{R}^{d_{model}}$.

### 2.2.4 Static covariate encoders

This encoder receives the static covariate vector $\tilde{\theta}_{\text{static}}$ via the variable embedding module as input and encodes it as conditional input information for different parts of the TFT, such as $c_v$ used by the variable embedding module for historical/future covariates, the starting cell state $c_c$ and hidden state $c_h$ of the LSTM in the Seq2Seq layer, and $c_e$ in the static enrichment layer. These state variables are calculated as

$$\left[ c_v^T, c_c^T, c_h^T, c_e^T \right]^T = W_{mlp} \tilde{\theta}_{\text{static}} \in \mathbb{R}^{(4 \times d_{model})} \qquad (3)$$

where $\boldsymbol{W}_{\text{mlp}} \in \mathbb{R}^{(4 \times d_{\text{model}}) \times d_{\text{model}}}$ is the learnable weight. Compared with the original TFT model which needs to pass through four separate GRN modules with different weights and optimize them separately in parameter learning, this paper only requires backward gradient propagation for a single fully connected neural network in this module, which improves the training speed.

### 2.2.5 Local and long-term temporal dynamics

To capture local features of the time series, such as important trend transitions or significant fluctuations, TFT uses a Seq2Seq encoding approach based on LSTM, which takes the embedded features of static covariates as initial weights, takes the features embedded in the covariates at each moment in history and future as input, and returns the encoding results for each moment simultaneously. The advantage of doing so is that it provides information on the relative positions of the time series elements, replacing the traditional fixed position encoding, and provides a reasonable inductive bias for each prediction step. For any moment $t+n$, where $n \in \{-k : \tau_{\max}\}$, the embedding vector of the input LSTM layer is $\tilde{\boldsymbol{\theta}}_{t+n}$ and the corresponding encoding vector for that moment is $\boldsymbol{\phi}_{t+n}$. Subsequently, this vector is passed through the gating module, LayerNorm module, residual connectivity module, and static enrichment layer to finally obtain the sequence of local feature vectors obtained by TFT encoding on the temporal data $\boldsymbol{\Psi} = \left[ \boldsymbol{\psi}_{t-k}, \boldsymbol{\psi}_{t-k+1}, \cdots, \boldsymbol{\psi}_{t+\tau_{\max}} \right]$. This process is calculated as follows

$$\tilde{\boldsymbol{\phi}}_{t+n} = \text{LayerNorm}\left( \tilde{\boldsymbol{\theta}}_{t+n} + \text{GLU}_{\tilde{\phi}}(\boldsymbol{\phi}_{t+n}) \right) \in \mathbb{R}^{d_{\text{model}}} \tag{4}$$

$$\boldsymbol{\psi}_{t+n} = \text{GRN}_{\psi}\left( \tilde{\boldsymbol{\phi}}_{t+n}, \boldsymbol{c}_e \right) \in \mathbb{R}^{d_{\text{model}}} \tag{5}$$

where $\text{GLU}_{\tilde{\phi}}(\cdot)$ and $\text{GRN}_{\psi}(\cdot)$ have different subscripts, implying the use of different weights; $\boldsymbol{c}_e$ provides the static covariate encoder with the static covariate condition information for the static enrichment layer. To obtain information on the correlation between different time steps of a time series over a wide field of view, TFT draws on the Transformer prototype based on a multi-headed attention mechanism to improve interpretability and long-term dependence in the prediction process. To enable a well-measured feature importance at each time step, TFT uses an attention calculation method with shared weights. Considering that $N = k + \tau_{\max} + 1$ is the total number of historical and future time steps, the number of heads of attention is $m_H$, and the tensor $\boldsymbol{\Psi} \in \mathbb{R}^{N \times d_{\text{model}}}$ has been obtained in the extraction of local temporal dynamic features, the corresponding query, key and value tensors for the head numbered $h$ are

$$\boldsymbol{Q}^{(h)} = \boldsymbol{\Psi} \boldsymbol{W}_Q^{(h)} \in \mathbb{R}^{N \times d_{\text{model}}} \tag{6}$$

$$\boldsymbol{K}^{(h)} = \boldsymbol{\Psi} \boldsymbol{W}_K^{(h)} \in \mathbb{R}^{N \times d_{\text{model}}} \tag{7}$$

$$\boldsymbol{V} = \boldsymbol{\Psi} \boldsymbol{W}_V \in \mathbb{R}^{N \times d_{\text{model}}} \tag{8}$$

where $\boldsymbol{W}_Q^{(h)}, \boldsymbol{W}_K^{(h)} \in \mathbb{R}^{d_{\text{model}} \times d_{\text{model}}}$ is a learnable weight and is different for each head; $\boldsymbol{W}_V \in \mathbb{R}^{d_{\text{model}} \times d_{\text{model}}}$ is also a learnable weight, but its weight is shared by all heads. For a single head, its self-attention matrix $\boldsymbol{H}^{(h)}$ is computed as

$$H^{(h)} = \text{Attention}\left(Q^{(h)}, K^{(h)}, V\right)$$
$$= \underset{\text{dim}=-1}{\text{Softmax}}\left(Q^{(h)} K^{(h)\text{T}} / \sqrt{d_{\text{model}}}\right) V \quad (9)$$
$$\in \mathbb{R}^{N \times d_{\text{model}}}$$

If all heads are considered, the average attention matrix $\tilde{H}$ is obtained as

$$\tilde{H} = \frac{1}{m_H} \sum_{h=1}^{m_H} H^{(h)} \in \mathbb{R}^{N \times d_{\text{model}}} \quad (10)$$

In the attention matrix $\tilde{H}$, each row represents the encoding result at the corresponding moment, i.e., a vector of dimension size as $d_{\text{model}}$. The vectors numbered $[t+1:t+\tau_{\max}]$ are the initial encoding results for each future time step in the prediction step. These preliminary coding results are then fed into the final layers of the feedforward network module and ultimately predict the confidence range of the target variable corresponding to each future time step.

### 2.2.6 Interval estimation

For the prediction range, the vector at each moment after multi-headed self-attention encoding can be denoted as $\beta_{t+n}$, where $n \in \{1, 2, \cdots, \tau_{\max}\}$. After the last few layers of residual connectivity and GRU layers, the encoding of each prediction time step $\epsilon_{t+n}$ s obtained, which is calculated as

$$\tilde{\epsilon}_{t+n} = \text{LayerNorm}\left(\tilde{\phi}_{t+n} + \text{GLU}_{\tilde{\epsilon}}(\epsilon_{t+n})\right) \in \mathbb{R}^{d_{\text{model}}} \quad (11)$$

$$\epsilon_{t+n} = \text{GRN}_{\epsilon}(\delta_{t+n}) \in \mathbb{R}^{d_{\text{model}}} \quad (12)$$

$$\delta_{t+n} = \text{LayerNorm}\left(\psi_{t+n} + \text{GLU}_{\delta}(\beta_{t+n})\right) \in \mathbb{R}^{d_{\text{model}}} \quad (13)$$

In this case, the different corner labels of GRN and GLU represent different network weights. Ultimately, the TFT's point prediction is based on the calculation of the prediction interval, which is achieved by simultaneously predicting various percentiles, such as the 10th, 50th and 90th at each time step using a linear decoder. The calculation is shown below

$$\hat{y}_{t+n}(q) = W_q \epsilon_{t+n} + b_q \quad (14)$$

where $\hat{y}_{t+n}(q)$ is the estimate of the target parameter $y$ at moment $t+n$ with time quantile $q$; $W_q$, $b_q$ are the weights and biases to be learned, respectively.

## 2.3  TFT oriented prognosis of LOCA

Based on the previous introduction of TFT, the structural features of the TFT prediction model mainly include gating mechanisms, variable embedding module, static covariate encoders, local and long-term temporal dynamics, and interval estimation. Compared with traditional prediction methods based on various deep learning models, such as classical LSTM or GRU, TFT has the following advantages for the prediction task of key parameters after LOCA:

1） The forgetting problem of models in dealing with long time series data is avoided. In order to capture the temporal characteristics of the data, prediction models were usually constructed in the past using recurrent neural networks. However, since the computation process is recursive, i.e., a new input is fed to the model at each moment, it leads to a change in the short-term memory, i.e., the activation state of neurons, within the model. The consequence of doing so is that the influence of earlier input data on the current or future model output is gradually reduced, i.e., the model forgets the earlier inputs, which in turn leads to larger prediction biases over time. Although there are novel gating-based LSTMs and GRUs that enable extended short-term memory, they do not completely solve the long-range dependence problem. TFT, based on the modeling of time-series data using LSTM, captures the interaction information of data points remotely using an improved multi-headed self-attention mechanism, so that the model has a global view and avoids the problem of forgetting early input data.

2） The efficiency of the use of valid information in the prediction task is improved. Since the data are fed into the TFT model at each time step to obtain aggregated coded features through feature embedding, this ensures robustness of covariate feature extraction at each time step. At the same time, these data are filtered by GRN with a gating mechanism on the features to achieve a certain degree of filtering of interfering data, such as noise or anomaly monitoring, and to highlight the contribution of key moments to prediction accuracy, such as monitoring important turning points or fluctuations in the data. In addition, the TFT model is able to receive input from multiple types of covariates, especially the support of static covariates represented by accident category and severity, enabling information transfer between the diagnostic step and the prognostic step after LOCA, bridging the information silos between the two phases.

3） The confidence level of the prediction task in safety critical scenarios is enhanced. Although the uncertainty calculation for the prediction step is implemented using only the simple idea of quantile regression, it provides the reactor operator with a more intuitive worst-case, best-case estimate of the system response than if there were only one output at a single time step. Uncertainty calculation is a critical function in scenarios where safety is a high requirement.

After constructing the TFT model containing the above description, this paper then proceeds to predict the key parameters after LOCA for the origination conditions and to evaluate and optimize the performance of the model. The workflow used in this paper is shown in Figure 6.

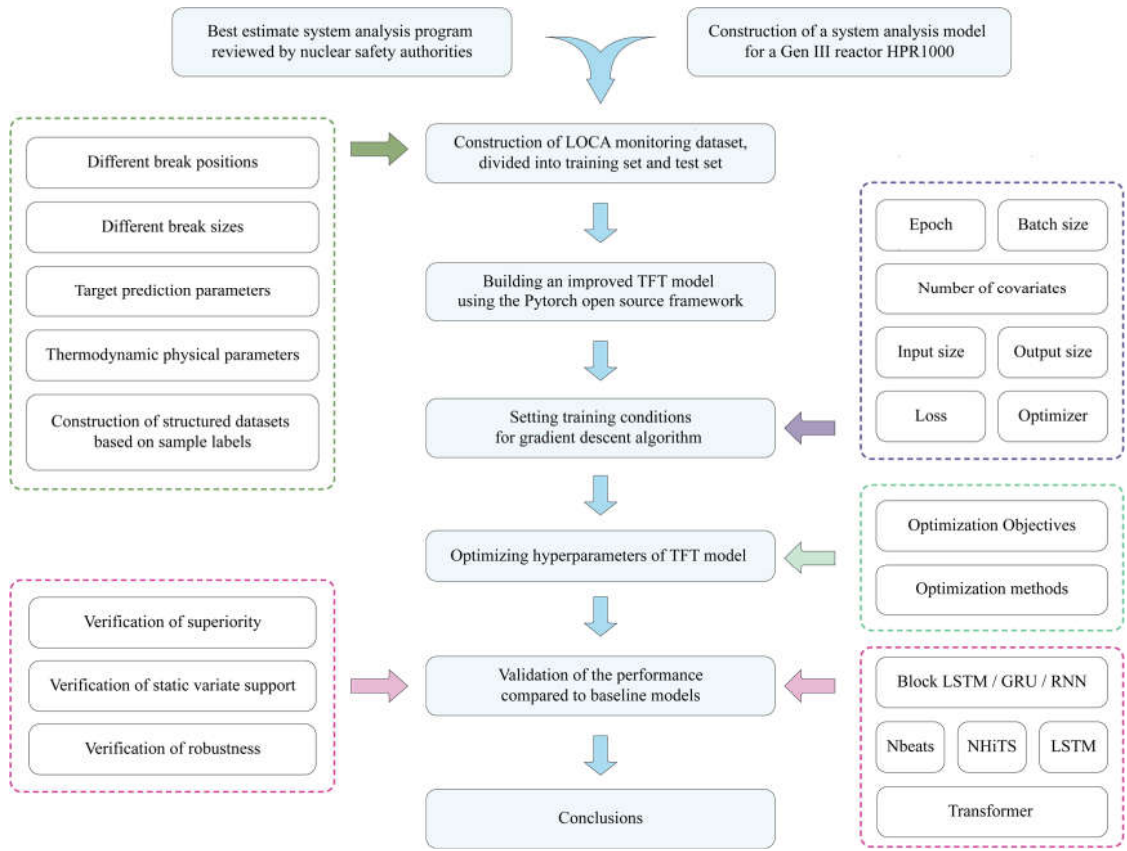

Figure 6 Flow chart of accident process prognosis after LOCA using TFT model

## 3. Data preprocessing and determination of hyperparameters

### 3.1 Data description and preprocessing

When a LOCA occurs in a reactor, the operator needs to go through five steps to complete the treatment of the incident: data collection, anomaly sensing, incident diagnosis, status prognosis, and finally mitigation decision [33]. Condition prognosis directly serves the subsequent mitigation decisions and is therefore a link directly related to the success or failure of incident management. Currently, operators use emergency operating procedures (EOPs) constructed during the reactor design phase in the event of a LOCA. However, the accident transients on which the EOP is based are only a few sparse design conditions with severe consequences, and cannot cover real-world scenarios that may occur. This can lead to a tendency for the operator to use the most conservative approach even in less severe transient situations, resulting in an unnecessary waste of resources, even if the conservative process does not effectively envelop the transient. Therefore, in order to construct an accident process prognosis method that can be used at different breach locations and breach sizes, a dataset for training the TFT model needs to be constructed. Considering that no real-world data on LOCA transients are available, it is necessary to simulate the LOCA process for different initiation conditions using suitable tools.

This paper uses a pressurized water reactor system analysis program, Advanced Reactor System Analysis Code (ARSAC), developed by the Nuclear Power Institute of China (NPIC). ARSAC is a modern pressurized water reactor transient analysis program that solves a non-equilibrium non-homogeneous Eulerian-Eulerian six-equation two-phase fluid model using a gas-liquid two-phase model framework. The development of ARSAC follows the standard six steps of program development, namely requirements analysis, physical model study, software design, coding, testing, verification and validation [17]. At present, ARSAC has been applied to several international benchmark problems, such as re-inundation experiments FLECHT-SEASET and other separation effect experiments, and large, middle and small breach loss of coolant accidents and other accident transient overall effect experiments, and the validation results show that the deviation of key parameters calculated by the ARSAC program from the experimental data is within a reasonable range. Therefore, the LOCA transients calculated using ARSAC are credible and reflect the reactor system response under realistic scenarios.

HPR1000 is a Gen III advanced nuclear power technology developed by China National Nuclear Corporation (CNNC) with a combined active and passive safety design concept. On the one hand, it is an evolutionary design based on the proven technology of existing pressurized water reactor nuclear power plants; on the other hand, it incorporates advanced design features, including 177 fuel assembly cores loaded with CF3 fuel assemblies, active and passive safety systems, comprehensive severe accident prevention and mitigation measures, enhanced protection against external events, and improved emergency response capabilities [3]. Some of the key technical parameters used in the simulation of different LOCA initiation conditions for HPR1000 are shown in Table 1. As well, at the moment when the transient occurs, the system analysis program runs at the steady state of the rated power.

Table 1 Some of the key technical parameters of HPR1000 at the onset of LOCA transients

| Parameter | Value |
| --- | --- |
| Power rating of the core | 3050MWt |
| Operating pressure | 15.5MPa |
| Height of cold active section of core | 3658mm |
| Average line power density | 173.8W/cm |
| Thermal design flow rate | $22840 \times 3m^3/h$ |
| Temperature of reactor PV inlet | 291.5℃ |
| Temperature of reactor PV outlet | 328.5℃ |
| Total volume of pressurizer | $51m^3$ |
| Design temperature of pressurizer | 360℃ |

In order to construct input cards that can be read by ARSAC, it is necessary to model the first loop and part of the second loop of the HPR1000, i.e. to illustrate the parameters of each reactor component and the connection relationships between the components, which is a process that can be represented in the form of a node diagram. The final node diagram of the HPR1000 reactor used to simulate the LOCA transient is shown in Figure 3, which contains the core of the reactor, the pressurizer, and key equipment on the three loops, such as the main pump and steam generator, as well as the main feedwater and steam co-tank of the second loop and the steam turbine for power generation.

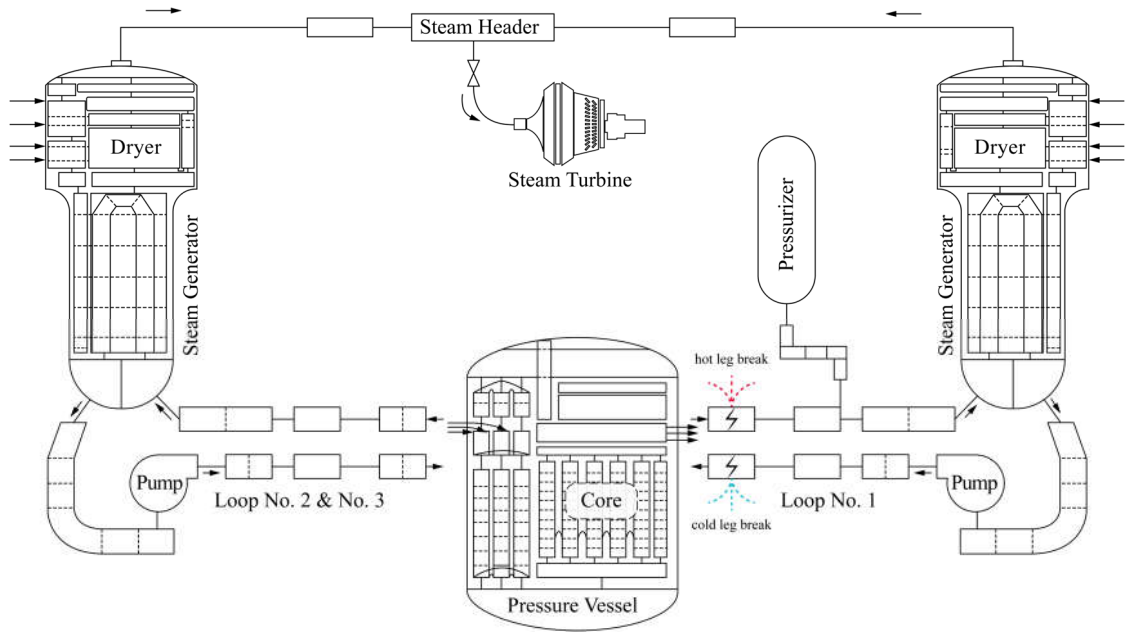

Figure 7 HPR1000 node diagram for simulating LOCA transients

In order to simulate different LOCA initiation conditions, two main settings were made: 1) firstly, it was determined that the cold leg breach occurred at the connection between the two nearest pipe nodes before the coolant inlet from the core in the first loop, and the hot leg breach occurred at the connection between the two nearest pipe nodes after the coolant exit from the core in the same loop; 2) secondly, the breach size started with an equivalent diameter equal to 0.1cm, and the step length is 0.2cm, and ends at an equivalent diameter of 35.5cm. The reason for setting different breach locations in the first place is mainly due to two considerations: 1) on the one hand, since there is always coolant flowing through the core compared to a hot leg breach, while a larger cold leg breach will result in a completely exposed core, the cold leg breach is the object of analysis in the reactor SAR, so the blind spot of the hot leg breach needs to be filled in the prognostic task for LOCA; 2) on the other hand, since the cold or hot leg is a category-based variable, it will help to extend the prognostic task to more accident category labels in the future. And the reason for setting more break sizes as an initiation condition is that compared to the large breaks where a dramatic system response occurs, small and middle breaks, although resulting in a less significant pressure relief process, still have the potential for complete core exposure, thus threatening the core integrity.

In the simulation of the LOCA cases, the simulation duration of each initiation event in this paper is 2000 seconds, and the state of the reactor at the current moment is recorded with a sampling frequency of twice per second. In the process of collecting covariates that contribute to the prediction of target parameters, this paper extracts a total of 30 directly monitored signals, 8 monitorable synthetic signals, and a fuel cladding maximum temperature parameter that is closely related to safety but not measurable, based on the physical signals that can be monitored by the actual instrumentation and control system of the HPR1000. The specific signal codes and the corresponding physical quantities for these parameters are shown in Table 2. The signal codes with the prefix "cntrlvar" are synthetic signals further calculated from several direct signals that reflect the reactor's operating state at the system level, and are therefore target parameters that can be applied to the prediction task.

Table 2 Correspondence of Signal codes and parameters that can be monitored by instrumentation and

control system

| Signal type | Signal code | Corresponding parameter |
| --- | --- | --- |
| Unmeasurable signal | cntrlvar_913 | Maximum core cladding temperature |
| Synthetic signals | cntrlvar_42 | Water level of pressurizer |
| | cntrlvar_121 | Mass flow rate of reactor coolant |
| | cntrlvar_11 | Water level of steam generator |
| | cntrlvar_100 | Maximum average temperature of loops |
| | cntrlvar_2 | Water level of PV |
| | cntrlvar_101 | Avg. temperature of the broken loop (1#) |
| | cntrlvar_102 | Avg. temperature of loop 2# |
| | cntrlvar_103 | Avg. temperature of loop 3# |
| Direct Signals | tempf_505010000 | Temperature of main feed water |
| | mflowj_505010000 | Mass flow rate of main feed water |
| | mflowj_566010000 | Mass flow rate of auxiliary feed water |
| | mflowj_537000000 | Mass flow rate of main steam |
| | p_540010000 | Pressure of steam line |
| | p_850010000 | Pressure of steam busbar |
| | voidf_811010000 | Water level of SI |
| | p_810010000 | Pressure of SI |
| | mflowj_811010000 | Mass flow rate of LHSI pump |
| | mflowj_806000000 | Mass flow rate of boron injection pump |
| | rktpow | Avg. power |
| | tempf_138010000 | Temperature of reactor core outlet |
| | tempf_155010000 | Temperature of the upper head |
| | p_155010000 | Pressure of reactor coolant |
| | p_260010000 | Pressure of pressurizer |
| | tempf_200010000 | Temperature of the broken loop (1#) hot leg |
| | tempf_300010000 | Temperature of hot leg of loop 2# |
| | tempf_400010000 | Temperature of hot leg of loop 3# |
| | tempf_250010000 | Temperature of the broken loop (1#) cold leg |
| | tempf_350010000 | Temperature of cold leg of loop 2# |
| | tempf_450010000 | Temperature of cold leg of loop 3# |
| | pmpvel_235 | Pump speed of the broken loop (1#) |
| | pmpvel_335 | Pump speed of loop 2# |
| | pmpvel_435 | Pump speed of loop 3# |
| | tempf_2700(1-5)0000 | Temperature of pressurizer surge tube (divided into 5 nodes) |
| | tempg_260010000 | Gas temperature of pressurizer |
| | tempf_262010000 | Liquid temperature of pressurizer |
| | tempg_281010000 | Upstream temperature of the safety valve of the pressurizer |
| | voidf_200010000 | Water level in the hot leg of the breakout loop |

Take the example of a middle break in the HPR1000 with a cold leg break size of 7.5 cm. Throughout the accident sequence, significant gas-liquid stratification of the reactor coolant system (RCS) occurs and there are two fuel temperature rises by gravity, which may lead to localized fuel damage. The first temperature rise is due to a loop water seal caused by the low point of the first loop system including the U-shaped elbow in front of the main pump and the lower part of the PV. This water seal causes the steam space to grow and the core to become exposed. When the break in the cold leg is exposed, the loop water seal at the low position of the reactor is removed, so the coolant is quickly re-entered into the core by the driving pressure head. The second temperature rise is caused by simple evaporation from the core. When the core is suddenly cooled due to the entry of new coolant, an early imbalance between the break flow and the safety injection flow can lead to a further drop in the PV water level, resulting in another bare core. The changes of the main synthetic monitoring signals after the occurrence of the middle break are shown in Figure 8. In the spray release phase, the PV water level signal decreases rapidly until the input of the position an injection box around 400 seconds; then, due to the formation of the loop water seal and the exposure of the core, the maximum temperature of the core envelope rises rapidly around 900 seconds; after 1600 seconds the loop water seal is lifted and the core water level rises again, at which time the maximum temperature of the envelope also starts to decrease.

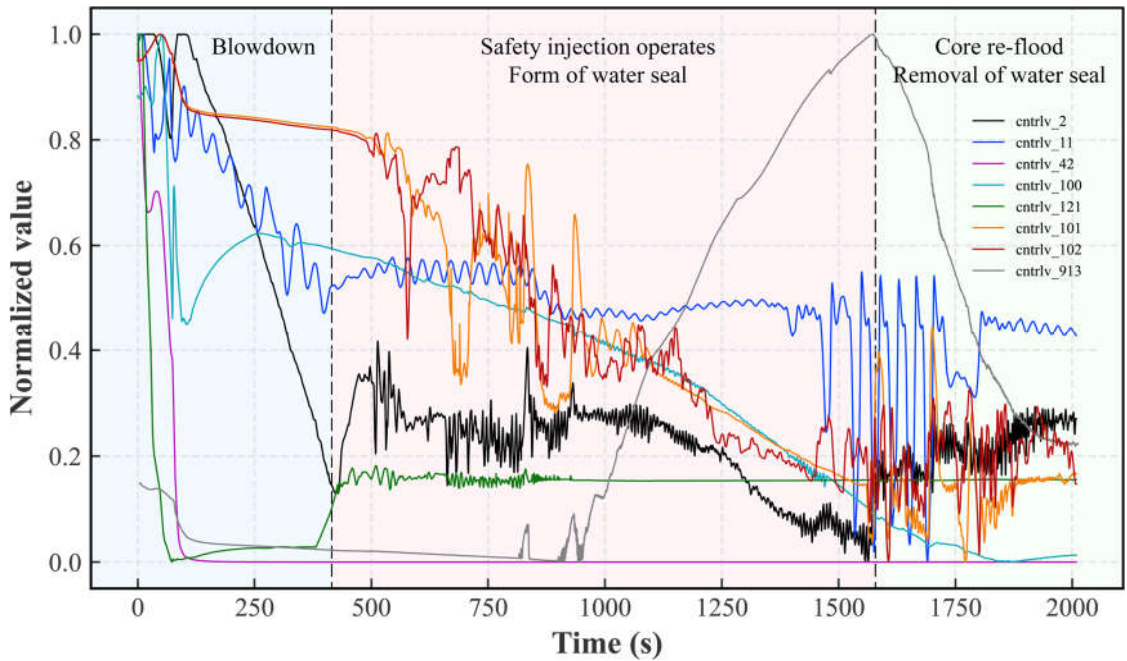

Figure 8 Variation of some key parameters in a typical cold leg middle break accident in 2000 seconds time

To reduce the computational complexity of the training process, the input to the TFT model can be simplified by reducing the number of covariates with high similarity. In this paper, the correlation between each pair of reactor signals is analyzed using Pearson's algorithm to obtain the signal correlation matrix including time and unmeasurable parameters, as shown in Figure 9. It can be seen that there are a large number of coefficients with correlations close to 1. Therefore, the coefficients are filtered by manual means. The trimmed coefficients are framed in red solid lines on the left side of the Figure 9. After signal screening, there are 13 direct monitoring signals left that can assist in the prediction of the target parameters.

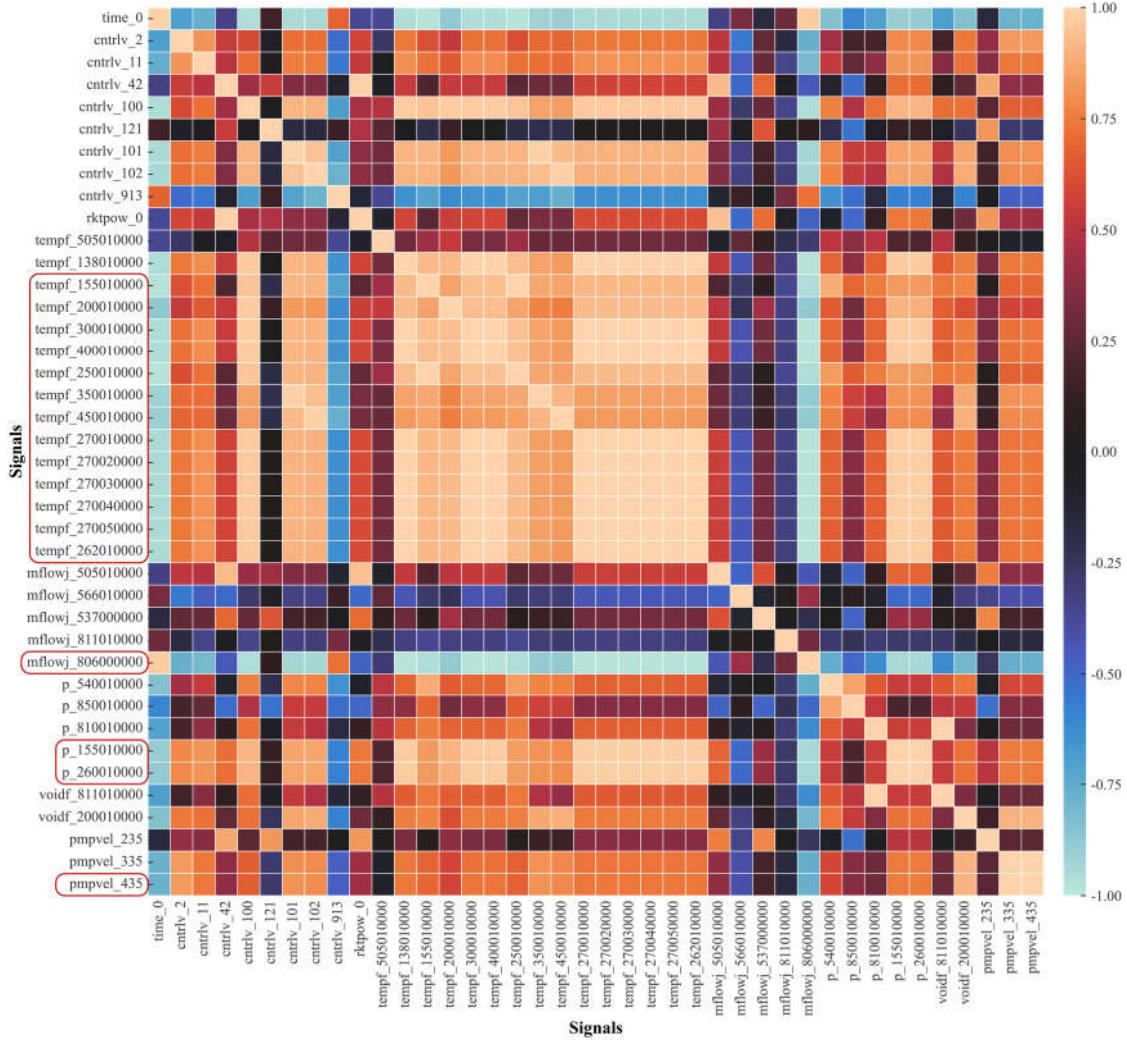

Figure 9 Heat map of Pearson correlation coefficients between pairs of reactor monitoring signals, and direct monitoring signals removed due to information redundancy

## 3.2 Training conditions

In order to train the TFT model, an explicit differentiable optimization objective, i.e., a loss function, is needed first. In order to meet the requirements of the TFT prediction model for interval estimation, it is necessary to use an aggregated quantile residual as a loss function and the aggregation is additive, so that it is calculated as

$$\mathcal{L}(\Omega, W) = \sum_{y_t \in \Omega} \sum_{q \in \mathcal{Q}} \sum_{\tau=1}^{\tau_{max}} \frac{\mathrm{QL}(y_t, \hat{y}_{t+\tau}(q), q)}{M \tau_{max}} \quad (15)$$

$$\mathrm{QL}(y, \hat{y}, q) = q \max(y - \hat{y}, 0) + (1-q) \max(\hat{y} - y, 0) \quad (16)$$

where $\Omega$ is the number of all time series samples in the training set; $\mathcal{Q}$ refers to the set of quartiles in the target, and the set used in this paper is $\mathcal{Q} = \{0.1, 0.5, 0.9\}$.

In the training process, the optimization method used is the Adam gradient descent optimizer. Adam has the advantages of the gradient descent algorithm with adaptive learning rate and the momentum gradient descent algorithm, which can improve the problem of prone to fall into the local minima of the loss function space while having a faster training speed. For the selection of the optimizer parameters, the initial learning rate is set to $lr = 1.0 \times 10^{-3}$, the coefficients used to calculate the running average of the gradient and its square are $\beta_1 = 0.9$ and $\beta_2 = 0.999$, the smoothing coefficient is $\epsilon = 1.0 \times 10^{-8}$, and the momentum decay coefficient is $4.0 \times 10^{-3}$.

In terms of data organization for the training process, all LOCA simulation databases are first partitioned into an 80% proportion of the training set and a 20% proportion of the test set. Then, the historical and prognostic data of each LOCA case are divided. The starting point of the prognosis is 100 seconds, which is due to the time consumed in order to undertake the diagnosis task of the accident [34], i.e., the initiating parameters of the transient are identified using the diagnostic model within 100 seconds of the transient occurrence, and at this point all known information is used to further predict the trend of the parameter of interest between 100 and 2000 seconds. Finally, in the actual training process, the small batch gradient descent method is used because if all the training data are input into the GPU memory at one time, it will lead to memory overflow; if only one sample is used to update the gradient at a time, i.e., random gradient descent, it will lead to a decrease in the convergence speed of the model. The data organization of all LOCA cases for the TFT model training and testing process is shown in Figure 10.

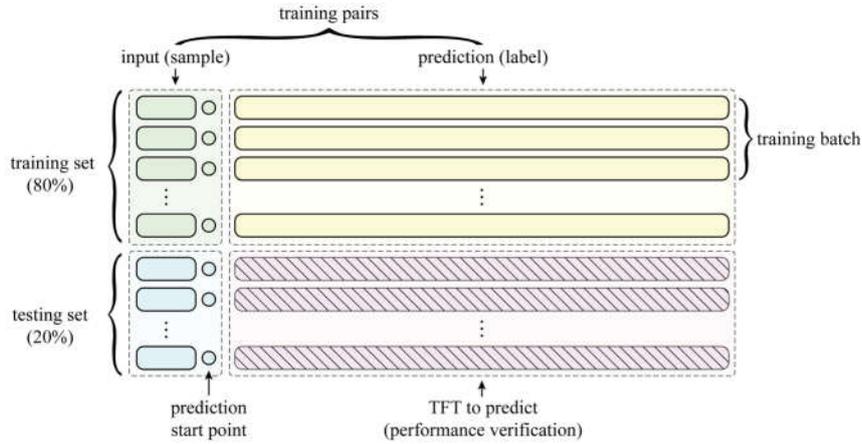

Figure 10 LOCA transient data organization for TFT model training and testing process

## 3.3 Determination of various hyperparameters

TFT is a deep neural network-based model, and therefore possesses numerous hyperparameters related to the structure of the network. Considering that it is impossible to optimize for each hyperparameter, the hyperparameters that are most likely to affect the prediction performance are selected as the objects to be optimized according to the characteristics of the core structure of the network. The hyperparameters to be optimized and their alternative parameter search ranges are shown in Table 3.

Table 3 The hyperparameters of the TFT model to be optimized and its search space

| Name of hyperparameters | Search space for parameter values |
|---|---|
| Hidden state size of the TFT ($d_{model}$) | $\{8, 9, 10, \cdots, 128\}$ |
| Number of attention heads ($m_H$) | $\{1, 2, 3, \cdots, 16\}$ |
| Number of layers for the LSTM | $\{1, 2, 3, \cdots, 16\}$ |

| Query to apply multi-head attention | {"future and past","only future"} |

The goal of hyperparameter selection was to confirm that the training speed and fitting ability of the model were sufficiently guaranteed during the formal experiments, so that optimization of the model's hyperparameters on the complete data set was not required, and therefore the performance of the model was only evaluated on individual samples (10 randomly selected LOCA cases with break diameters between 6.5 and 10.5 cm) and the appropriate hyperparameters were selected on this basis. In combination with the actual performance requirements of the post-LOCA parametric prediction problem, i.e., high requirements in terms of accuracy and confidence, two indicators need to be focused on when selecting the hyperparameters: 1) the residual distribution between the 50% quantile of the predicted and true measurements of the parameters has a mean and variance as close to zero as possible; 2) the true values of the parameters are as much as possible enveloped in the 10% and 90% quantile of the predicted values between the predicted and 90% quartiles. Therefore, the optimization problem with hyperparameters can be defined as

$$\begin{aligned}
\min \quad & f_{\mu,case}(\boldsymbol{x}) + f_{std,case}(\boldsymbol{x}) + f_{pct,case}(\boldsymbol{x}) \\
\text{s.t.} \quad & x_1 = d_{\text{model}} \in \{8, 9, 10, \cdots, 128\} \\
& x_2 = m_H \in \{1, 2, 3, \cdots, 16\} \\
& x_3 = LSTM\_layers \in \{1, 2, 3, \cdots, 16\} \\
& x_4 = Full\_attention \in \{True, False\}
\end{aligned} \quad (17)$$

where $f_{(\cdot),case}(\boldsymbol{x})$ represents the calculated function of the metric for a certain performance of the prediction process under the hyperparameter condition $\boldsymbol{x}$; $\mu$ and $std$ imply the mean and variance of the normal distribution when the residuals are fitted with a normal distribution, respectively; and $pct$ refers to the proportion of the prediction interval in which the true monitored value exceeds the predicted value by more than 10% and 90% of the quantile. In order to improve the speed of hyperparameter optimization, the performance metrics of the model are calculated after 100 training epochs. Although the optimization problem has three objective functions to be optimized, the single objective optimization problem constructed by summing them additively is sufficient to meet the performance requirements of this work.

In this paper, we use a Bayesian optimization (BO) approach called Expected Improvement (EI) [35] for the selection of hyperparameters. BO consists of two main elements: 1) the first component is a probabilistic agent model, which consists of a prior distribution and an observation model describing the data generation mechanism, such as a Gaussian process or a probabilistic tree model, and the observation model used in this paper is the Probabilistic Random Forest model; 2) the second component is an optimization objective, which describes a sequence of sampling and query processes to the best extent. The implementation algorithm for the single-objective optimization object of this paper is shown in **Algorithm 1**. The optimization iteration of hyperparameters is chosen to be 100 times, and the convergence curve of the optimization process is shown in Figure 11. Finally, the hyperparameters of the TFT network used for training are obtained as $d_{\text{model}} = 123$, $m_H = 11$, $LSTM\_layers = 15$ and $full\_attention = False$.

| **Algorithm 1: Bayesian optimization (single target)** |
|---|
| Input: $\boldsymbol{\theta}_0$ as the starting point of hyper-parameter; $\boldsymbol{\Theta}$ as hyper-parameter space; $\mathcal{D}_0$ as a container to collect observed trajectories; $n_{\max}$ as the maximum number of iterations; $\alpha_0 : \boldsymbol{\theta}_0 \to \mathbb{R}$ as a probabilistic surrogate function based on observation sequence |

```
1:   for n in {0,1,2,⋯,n_max − 1} :
2:        select new θ_{n+1} = arg min_θ α_n(θ_n; 𝒟_n) ∈ Θ
3:        obtain new observation y_{n+1} based on θ_{n+1}
4:        update trajectory container 𝒟_{n+1} = {𝒟_n, (θ_{n+1}, y_{n+1})}
5:        update surrogate model α_n --𝒟_{n+1}--> α_{n+1} based on posterior inspection
6:   end
     Output: θ_opt corresponding to y_opt = min(y_1, y_2, ⋯, y_{n_max})
```

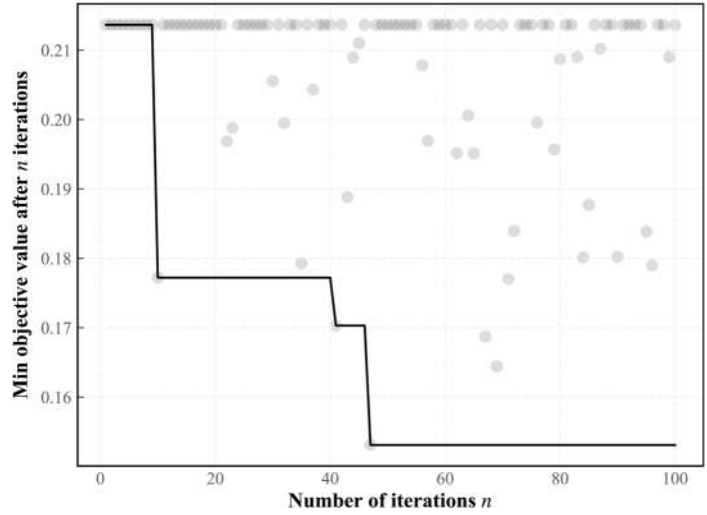

Figure 11 Convergence curves in hyperparameter optimization process

## 4. Prediction results and discussion

### *4.1  Verification of superiority*

In order to evaluate the performance of the TFT-based post-LOCA critical parameter prediction model for reactors proposed in this paper, several benchmark models and advanced deep learning models need to be used for comparison. Considering that the models are required to have a strong prediction accuracy and a confidence range that can be estimated in the post-LOCA prediction scenario, the following principles are followed in the selection of the comparison models:

（1） The prediction model is global rather than local. This means that the model can be tested on the training set and then can be directly inferred on the test set without further optimization of the model prior to inference [36].

（2） The model is able to perform estimation of confidence intervals. That is, it is able to perform uncertainty estimation for the computational prediction step as the TFT model used in this paper.

（3） The model is capable of receiving historical covariates as input. Although the TFT model used in this paper is capable of receiving static, historical, and future covariates as inputs, the restriction on the types of covariates supported by the comparison model is relaxed, considering that historical covariates may contribute the majority of information in the forecasting process.

Considering the above requirements, the comparison models used in this paper contain NiHiTS [37], Nbeats [38], Transformer [39], LSTM and Block-LSTM, GRU and Block-GRU, RNN and Block-RNN. The three neural network prediction models with "Block" prefixes are unique compared to the prefix-less models in that they use a fully connected network to produce a fixed-length output after encoding a fixed-length input block using a recurrent encoder, and therefore have a faster prediction speed.

In this paper, two synthetic monitoring parameters highly relevant to system safety are selected for prognosis: the PV water level with number cntrlvar_2, and the average temperature of the coolant in the breakout loop with number cntrlvar_101. After training the TFT model using randomly selected time-series data as shown in Figure 10 and testing on some of the remaining data, the results of the predicted PV water level parameters under hot and cold leg LOCA are obtained as shown in Figure 12 and Figure 13, respectively; the prediction results of the average temperature of the break loop under hot and cold leg LOCA are shown in Figure 15 and Figure 16, respectively. Overall, although the prediction results using TFT have different degrees of lags at key turning points and locations of drastic changes, the confidence intervals are basically able to envelop the true parameter changes, indicating a high degree of confidence in the prediction results. In addition, the distributions of the residuals between the 50% quantile of the predicted and true simulated values of the two parameters are shown in Figure 14 and Figure 17, respectively. It can be seen that the residual variables roughly follow a Gaussian distribution and have a mean and variance close to zero, thus reflecting the high accuracy of the prognosis.

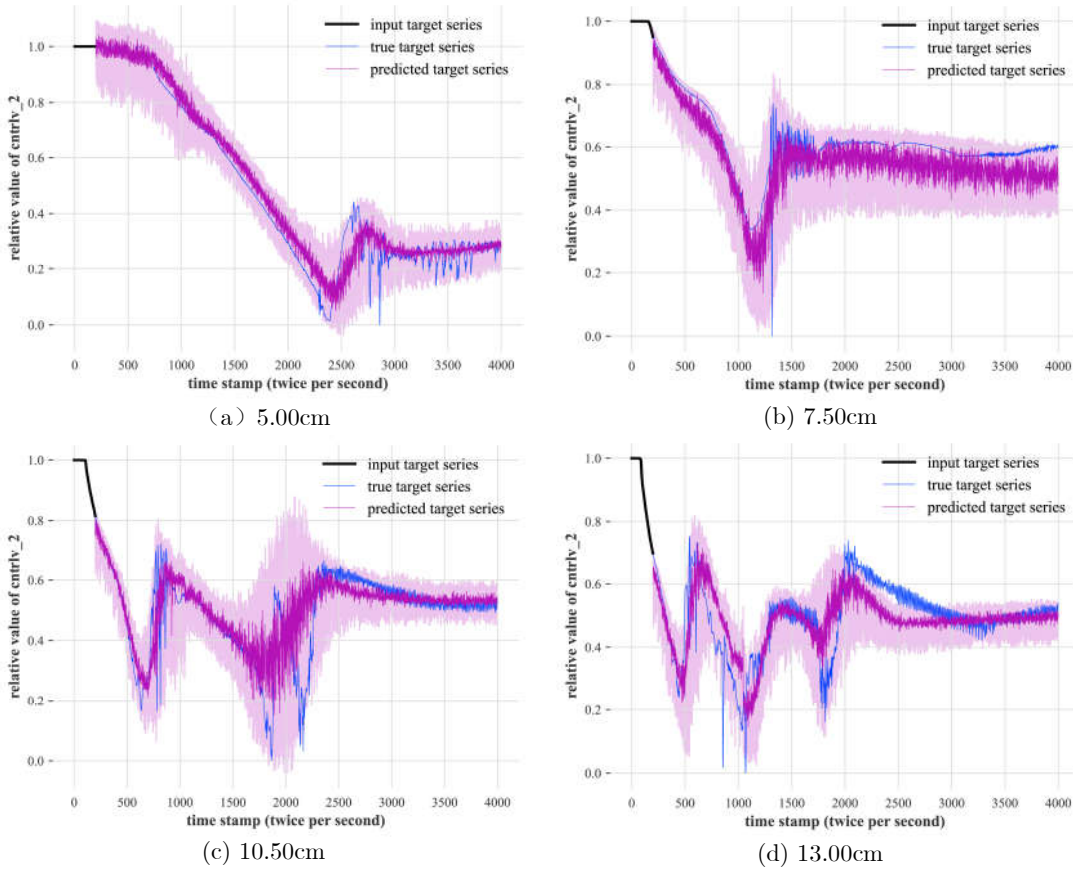

(a) 5.00cm  (b) 7.50cm
(c) 10.50cm  (d) 13.00cm

Figure 12 Samples of PV water level prognosis under hot leg LOCA

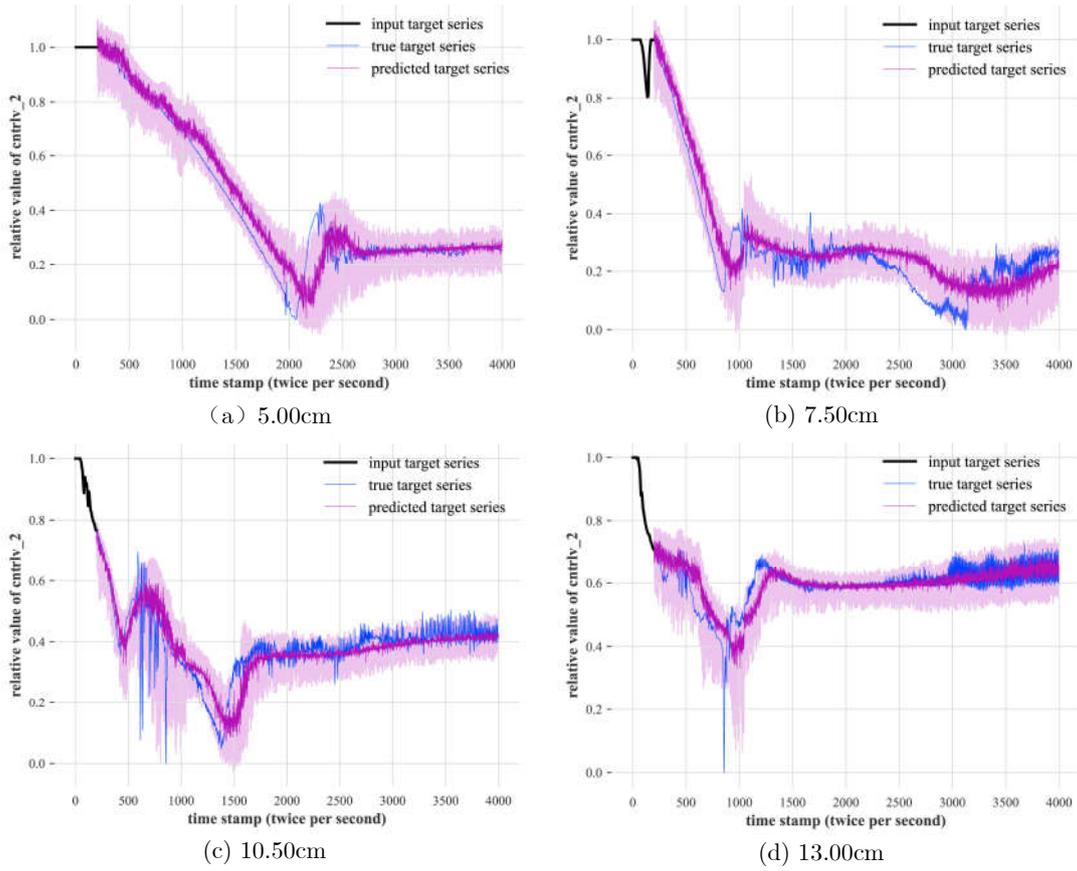

(a) 5.00cm  (b) 7.50cm

(c) 10.50cm  (d) 13.00cm

Figure 13 Samples of PV water level prognosis under cold leg LOCA

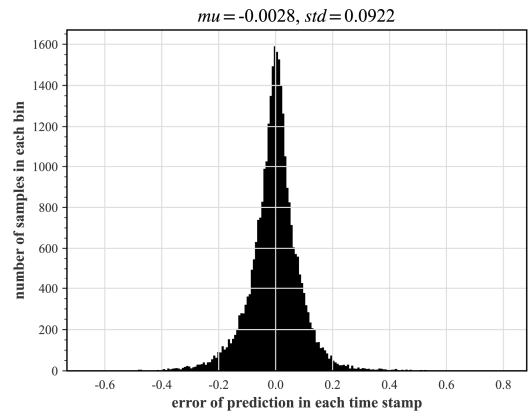

Figure 14 Distribution of PV water level prediction 50% quantile deviation from measured value

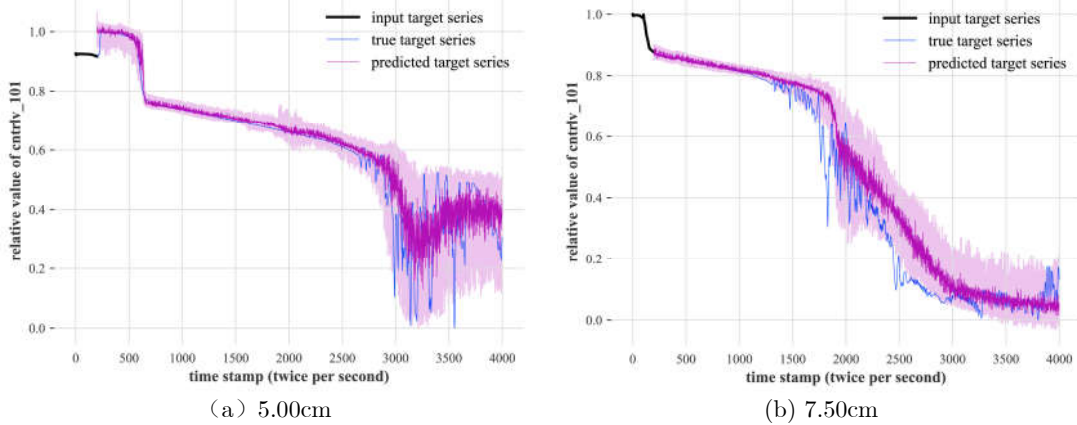

(a) 5.00cm  (b) 7.50cm

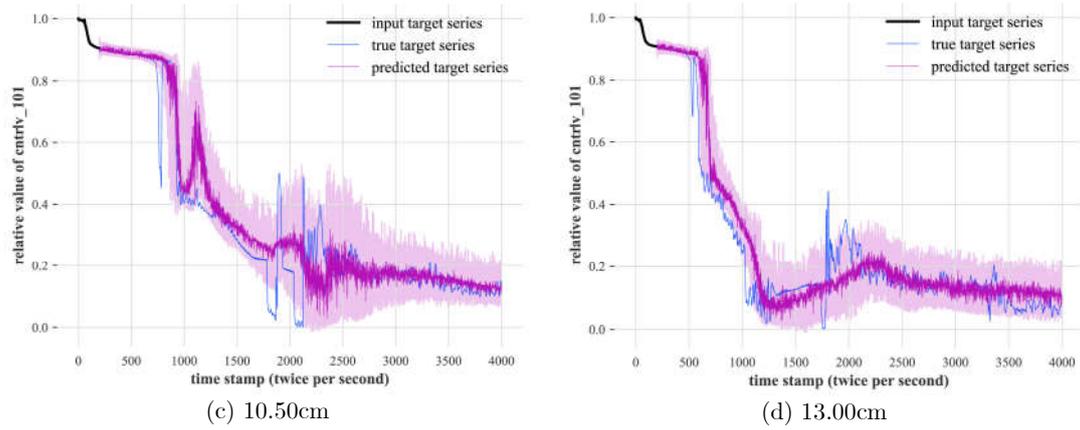

(c) 10.50cm

(d) 13.00cm

Figure 15 Samples of average temperature prognosis of the breaking loop under hot leg LOCA

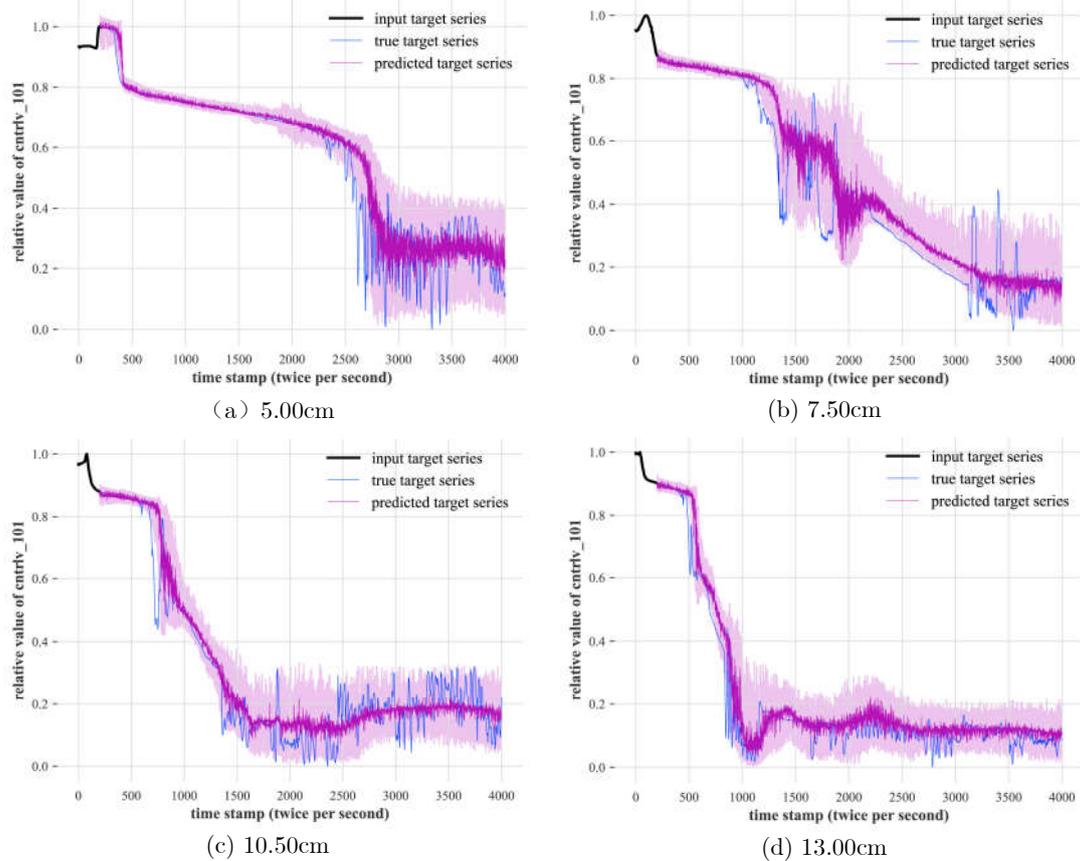

(a) 5.00cm

(b) 7.50cm

(c) 10.50cm

(d) 13.00cm

Figure 16 Samples of average temperature prognosis of the breaking loop under cold leg LOCA

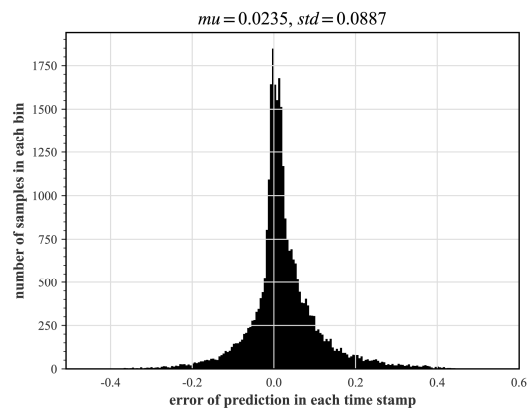

Figure 17 Distribution of deviations of the 50% quantile of the predicted value from the average

temperature measurement of the breaking loop

After obtaining the prediction results of the TFT model, the other prediction models used for comparison were trained with the same training method and the PV water level and the average temperature of the breach loop were predicted separately to obtain a comparative performance index of different prediction methods, and the performance pairs are shown in Table 4. Among the six specific evaluation metrics selected, the TFT model used in this paper obtained the highest performance in four of them. Therefore, it can be shown that the TFT model used in this paper is significantly superior for the task of prognosis of reactor accident parameters.

Table 4 Performance comparison of TFT models and different prognostic methods

|  | Mean value of error distribution | | Variance of the error distribution | | Proportion of measured values within the confidence range | |
| --- | --- | --- | --- | --- | --- | --- |
|  | PV water level | Average temperature of the breaking loop | PV water level | Average temperature of the breaking loop | PV water level | Average temperature of the breaking loop |
| TFT | **-0.0028** (1st) | **0.0235** (6th) | **0.0922** (1st) | **0.0887** (1st) | **91.21%** (1st) | **87.63%** (3rd) |
| Transformer | 0.0053 | *0.0039 (1st)* | 0.1154 | 0.1259 | 89.60% | 85.27% |
| NHiTS | 0.0045 | 0.0077 | 0.1392 | 0.1266 | 89.42% | 86.08% |
| Nbeats | -0.0973 | -0.1172 | 0.1856 | 0.2124 | 84.67% | 81.56% |
| LSTM | 0.1456 | 0.1787 | 0.2502 | 0.3091 | 74.04% | 79.24% |
| GRU | 0.1968 | 0.2085 | 0.2579 | 0.2965 | 73.39% | 73.46% |
| RNN | 0.3071 | 0.2942 | 0.2950 | 0.2367 | 70.53% | 68.32% |
| Block-LSTM | 0.0032 | 0.0046 | 0.1421 | 0.1368 | 90.21% | *89.74% (1st)* |
| Block-GRU | -0.0043 | 0.0060 | 0.1583 | 0.1344 | 88.70% | 89.69% |
| Block-RNN | 0.0089 | -0.0186 | 0.1927 | 0.1496 | 85.32% | 87.54% |

## 4.2 Verification of robustness

In the case of LOCA, the strong pressure drop and the spray release of the gas-liquid two-phase flow can cause a certain degree of rheological vibration in the first loop system and interfere with the accuracy of measurement of physical quantities at each measurement point, so it is necessary to evaluate the model prognostic capability in the case of different degrees of noise disturbance. The object used for the evaluation process is the TFT model trained in Table 4, which relies on the training data as a result of the simulation of LOCA by the system analysis program without additional added noise. Since monitoring data from real LOCA scenarios are not available, the deviation distribution of model predictions after adding noise with different signal-to-noise ratios (SNR) to the sequence of historical target parameters and the sequence of historical covariates on which the model predictions depend will be analyzed. In this paper, we consider the case where the SNR levels are $[40.0, 30.0, 25.0, 20.0, 15.0]$ and ignore the uncertainty of the noise on the upstream diagnostic task, i.e., the static covariates (break size and break location) included in the target parameters do not change. The reason for not considering the uncertainty of static covariate labels is that a transient representation-based diagnosis method tolerant to noise has been proposed by the authors of this paper in [34], which is able to extract valid accident representations and perform high accuracy diagnosis from monitoring data containing a mixture of crippled data and strong noise The two parameters that are the object of analysis are the PV liquid level signal of the reactor and the average temperature of the breaking loop.

The parameters of the bias distribution, i.e., the mean and variance, of the output of the prediction model using different SNRs are obtained with no changes to the static covariates and time variables, as shown in Figure 18. It can be seen that with the increase of noise, the variance of the prediction error of the water level signal and temperature signal does not show much change, and the absolute value of the prediction error of the temperature signal does not show large fluctuations. The only change that is more obvious is that the absolute value of the prediction error of the water level signal has a large increase with the increase of the SNR. After analysis, this is due to the existence of a wide range of low-frequency oscillation data characteristics of the water level in the middle LOCA range (e.g., Figure 12 (c,d)), resulting in the TFT model in predicting the signal changes in this interval will pay more attention to the relative position information of the data points, making the predicted initial value more sensitive to the mean value of the error. Specifically seen is the predicted performance of the TFT for PV water level for size 10.5 cm hot leg LOCA at SNRs of 40.0 and 15.0, respectively, as shown in Figure 19.

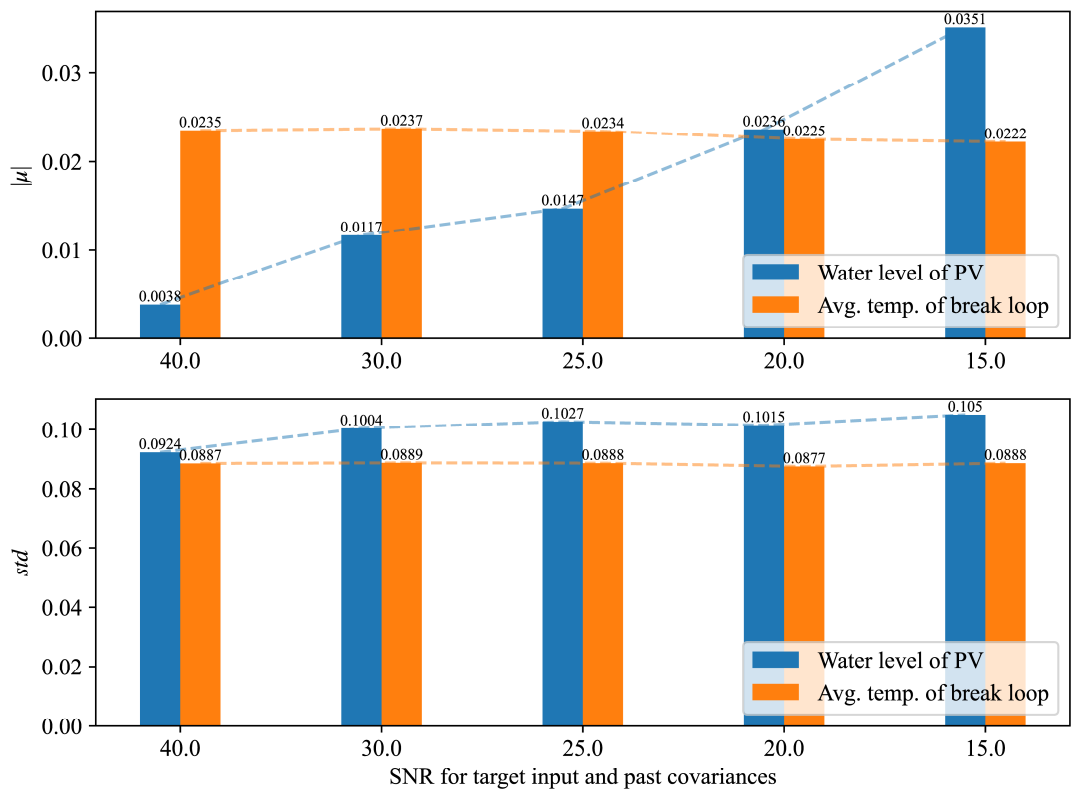

Figure 18 Prognostic performance of TFT models on test sets with different SNRs for input parameters

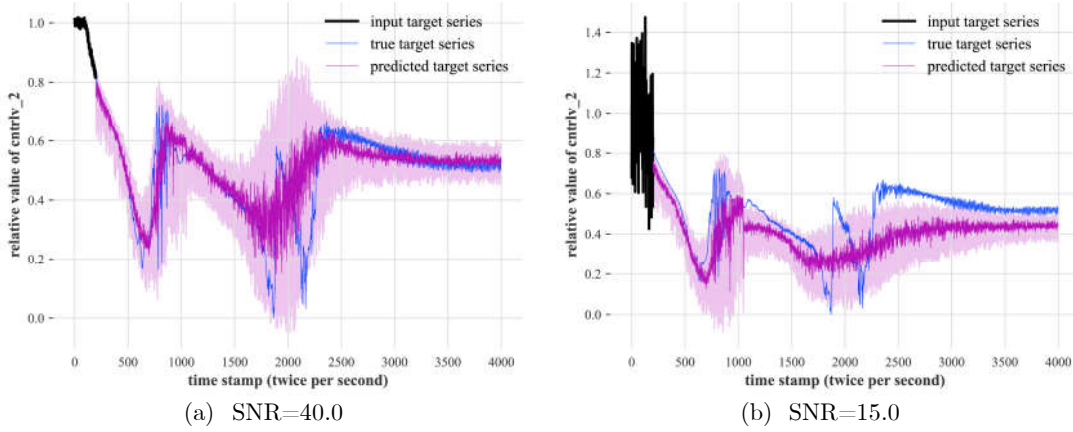

(a) SNR=40.0  (b) SNR=15.0

Figure 19 The performance of TFT model for PV water level prediction at SNR of 40.0 and 15.0

## 4.3  Verification of static variate support

To explain the outstanding performance of the TFT model under disturbed data, we make the following speculations: 1) on the one hand, the TFT model is able to understand the effective features of the accident from the disturbed input data; 2) on the other hand, the static covariates of the data received by the TFT already contain sufficient information about the accident features. To demonstrate this, this paper next conducts ablation experiments on the static covariates in the data input, i.e., the prediction performance of the TFT model at different SNRs after losing the information on the location of the break and the size of the break. The experimental scheme consistent with the above section is used here to obtain the model prediction performance at each SNR value, as shown in Figure 20.

As can be seen in the figure, with increasing noise levels, there is a significant increase in both the absolute value of the mean and variance of the prediction value errors. This is due to the fact that with the loss of static covariates, the information used to aid in prediction is provided only by the remaining monitorable historical covariates, and therefore there is a decline in the ability of the TFT model to obtain valid characteristics about the accident in the presence of high noise, which is in line with our first. It can be seen that data with accurate historical covariate information, i.e., with correct accident diagnosis before prediction, can guarantee the accuracy of prediction results in a high noise environment, as shown in the comparison of Figure 20 and Figure 19, which is consistent with our second speculation. In summary, the TFT model used in this paper can support static covariates, a feature that can more robustly assist operators in prognosing accident processes.

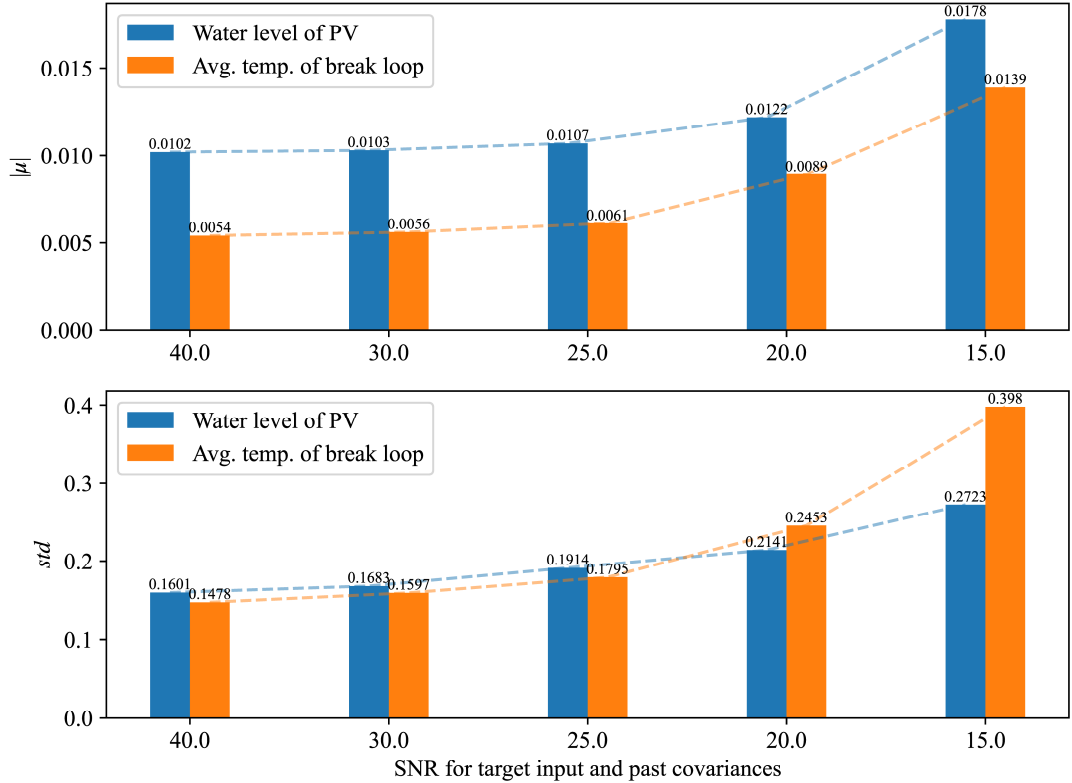

Figure 20 Prognostic performance of TFT models on the test set with different SNRs for input parameters under the condition of no static covariates

## 5. Conclusion

In this paper, an advanced composite deep learning model, Temporal Fusion Transformer (TFT), is applied for the first time to the task of prognosing critical system safety level parameters after loss of coolant accidents in reactors. The model used in this paper has three advantages: 1) the structural superiority of TFT guarantees the accuracy of prediction; 2) the support of historical covariates, future covariates, and static covariates by the prediction method improves the efficiency of data usage; 3) the estimation of upper and lower bounds of parameter values at a single time step by the model algorithm can effectively enhance the confidence of the prediction results to the reactor operators.

The data used in this paper were obtained from ARSAC, the best estimation program trusted by the Chinese nuclear energy authorities for reactor system analysis. Before formal training of the TFT model, a black-box optimization algorithm based on Bayesian theory was employed in this paper to determine the hyperparameters of the model. Subsequently, a randomly selected dataset is used as input, and the completed global TFT model is trained and applied to the prediction task on the test set. The prognostic performance of the TFT model is used to compare with several advanced deep learning-based prediction methods and demonstrates the superiority of the method in terms of prediction accuracy and confidence. In addition, the predictive capability of the TFT model was tested with several inputs of different levels of noise and it was found that although there is a larger effect on the predicted deviation of the core water level as the signal-to-noise ratio (SNR) increases, there is no greater effect on the standard deviation of the predicted error variables. Finally, by analyzing the forecasting performance of the TFT model with ablated static covariates at different levels of SNR, it is demonstrated that the tolerance of the TFT model to high-intensity noise lies in two parts: on the one hand, the model is able to understand its representation of the initiating event from the static covariates; on the other hand, the model is able to extract valid accident representations from directly monitorable parameters as historical covariates.

In conclusion, the TFT model used in this paper has a more precise and trustworthy ability to prognose the process of reactor loss of coolant accident. More importantly, the work in this paper provides a reliable and robust basis for reactor post-accident disposal decisions, and takes over the effective information for accident diagnosis, making the chain of post-accident reactor system maintenance more integral and smoother. Thus, the work in this paper makes a positive contribution to the establishment of a more intelligent and lightly burdened approach to reactor system maintenance.